\documentclass{llncs}
\usepackage{times,epsfig,graphicx}
\usepackage{algorithm,algorithmic}
\usepackage[usenames]{color}
\definecolor{MyDarkBlue}{rgb}{0,0.08,0.45}

\newcommand{\hide}[1]{}

\newcommand{\bfx}{{\bf x}}
\newcommand{\bfy}{{\bf y}}
\newcommand{\bfw}{{\bf w}}

\newcommand{\bfv}{{\bf v}}

\newcommand{\bfz}{{\bf z}}

\newcommand{\al}{{\alpha}}

\begin{document}

\title{Multi-task learning of time series and its application to the travel demand}
\author{Boris Chidlovskii}

\institute{Naver Labs Europe \\
6 chemin Maupertuis, 38240 Meylan, France\\
\texttt{boris.chidlovskii@naverlabs.com}}

\title{Multi-task learning of time series and its application to the travel demand}

\maketitle

\begin{abstract}
We address the problem of modeling and prediction of a set of temporal events in the context of intelligent transportation systems. To leverage the information shared by different events, we propose a multi-task learning framework. We develop a support vector regression model for joint learning of mutually dependent time series. It is the regularization-based multi-task learning previously developed for the classification case and extended to time series. We discuss the relatedness of observed time series and first deploy the dynamic time warping distance measure to identify groups of similar series. Then we take into account both time and scale warping and propose to align multiple time series by inferring their common latent representation. We test the proposed models on the problem of travel demand prediction in Nancy (France) public transport system and analyze the benefits of multi-task learning. 
\end{abstract}

\section{Introduction}
\label{sec:intro}

Time-dependent entities and their analysis represent a great challenge in multiple domains, including finance, science, medicine, entertainment, etc. In the domain of transport, intelligent transportation systems have been designed to track and analyze a large number of time-dependent events, such as vehicle position, road load, travel time and demand, service reliability, traffic density, etc. Efficient planning and management of transportation network, in particular, an adequate response to changing traffic conditions, require an accurate modeling and real-time prediction of all these time-dependent entities.
 
\begin{figure}[ht]
\centering{
\includegraphics[width=10cm]{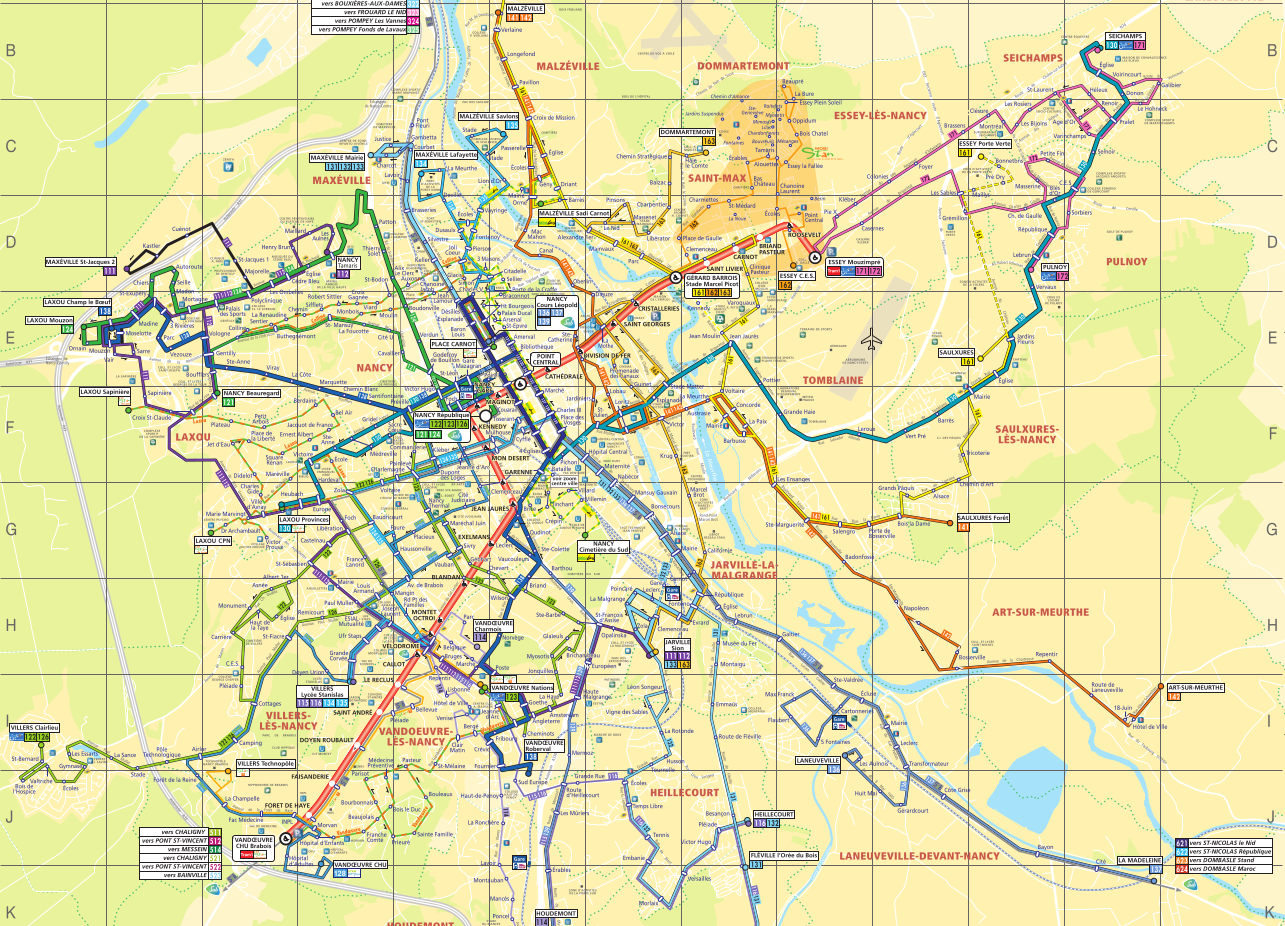}} 
\caption{Nancy public transport network.} 
\label{fig:nancymap}
\end{figure}

\begin{figure}[ht]
\centering{
\includegraphics[width=8.5cm]{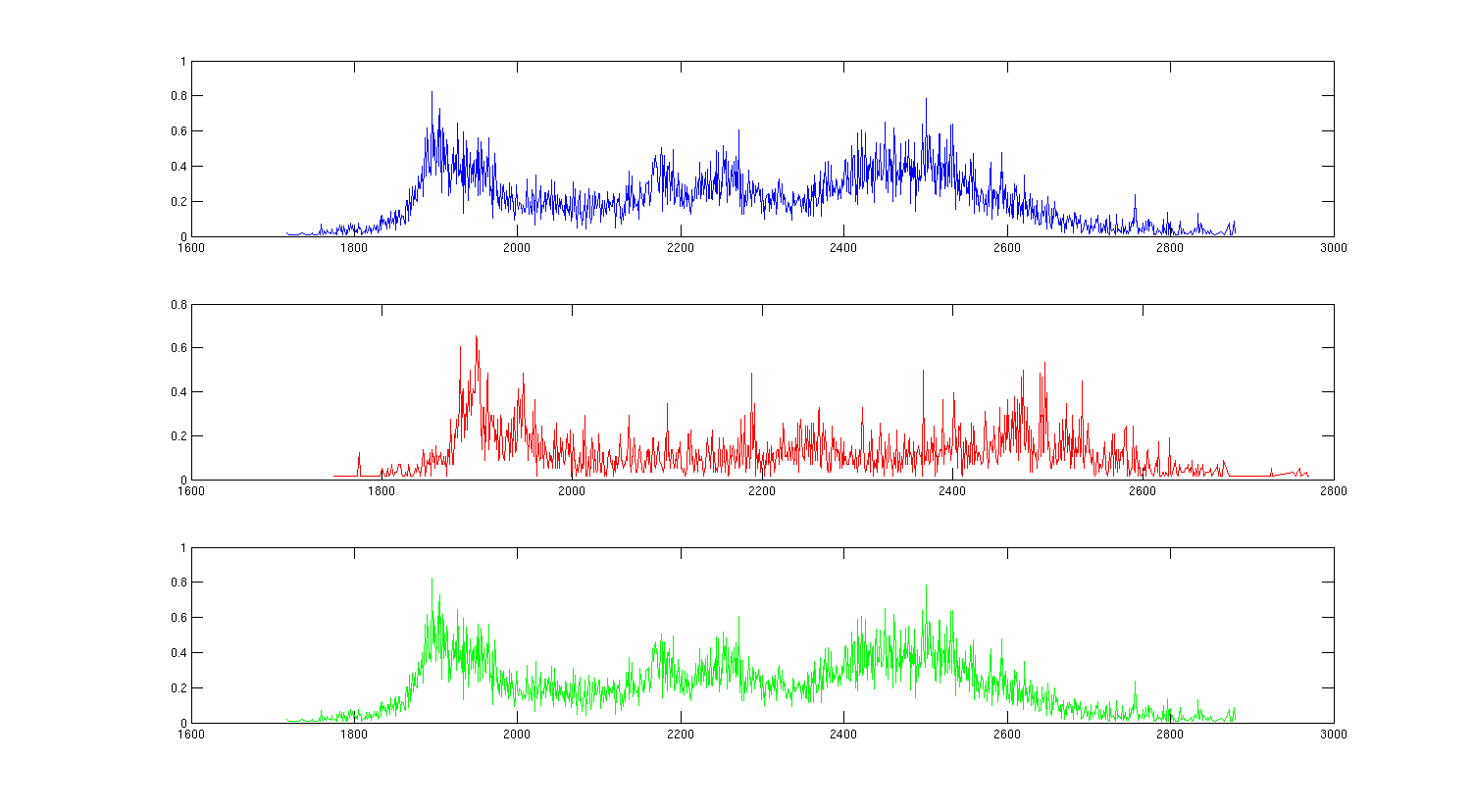}}
\caption{Three daily travel demand time series.} 
\label{fig:nancy-1day}
\end{figure}

It is common to represent time-dependent entities as time series. In this paper we address the problem of simultaneous modeling and prediction for a set of time series. Our work is motivated by the problem of travel demand prediction in a public transport system. Figure~\ref{fig:nancymap} shows the public transport network of Nancy city (France), Figure~\ref{fig:nancy-1day} reports three time series representing the daily passenger travel demand in different network locations. In these series, one item refers to the number of passengers boarding public transport vehicles (buses and trams) during a certain period of time. Similar and dissimilar patterns are easily recognizable in the series, they are often relevant to peak and off-peak times.

In statistics and signal processing, autoregressive models have been intensively used to study time-varying processes. An autoregressive (AR) model specifies that the output variable depends linearly on its own previous values. Combined with the integrated moving average (IMA) it forms a more general ARIMA model of time series; the later can be further generalized to a SARIMA model that takes into account seasonal effects~\cite{mills90}. Multiple techniques developed by the machine learning community, such as neural networks, perceptrons, support-vector machines and regressions, have been also adapted to model time series~\cite{sapankevych09}. 

In the transportation domain, the time series has been a subject of multiple studies in the context of traffic flow forecast~\cite{tan09}. The most recent study presents an extensive experimental protocol allowing to compare both autoregressive models and machine learning techniques, and to infer a general picture of the advantages and limitations of each approach~\cite{lippi13}.

In the most conventional setting, all time series are modeled individually. However, if they are related and share important information, a multi-task learning approach can be proposed to model all of them in a joint way. 

Complex dependencies between multiple time series is of high interest for anomaly detection~\cite{qiu2012,chawla2012}. Causality analysis discovers the temporal dependencies between a large number of time series data coming from different source, like sensors, logs or physical measurements. Inferring the Granger causality between series became particularly popular, for its scalability and interpretable insights on the possible reasons of anomalies~\cite{qiu2012}.

The {\it multi-task learning} is aimed at leveraging the information of multiple, mutually related learning tasks to make more accurate predictions for the individual tasks~\cite{caruana1997}. Related information contained in tasks can be exploited to mutually increase the quality of predictions. Examples of successful application domains for multi-task learning include computational biology, natural language processing, computer vision~\cite{leiva2012}, where multiple biological, textual and visual object classes may share some of the relevant features.

In the multi-task learning, the prediction accuracy in each task is leveraged by making use of data from the other tasks. Previous work has focused on relating the tasks via regularization~\cite{evgeniou2004}, mutualization~\cite{ecml12}, covariance sharing~\cite{bonilla07} and mainly addresses the classification case. In this paper we extend the regularization-based technique~\cite{evgeniou2004} to the time series.

Our motivation example for developing a multi-task learning approach is the travel demand prediction in a public transport network. We integrate a number of relevant features in prediction models. 

We first collect historical data of the travel demand in a public transport system. It is extracted from a massive collection of individual passenger boarding events in the network observed over some period of time.  
Second, we provide the calendar data, including weekends, school vacation, public holidays; these data are often critical to estimate the level of travel demand. Third, we integrate weather information, such as temperature, precipitation and wind speed, as factors that may also impact the demand. Lastly, we can benefit from the traffic information, describing the traffic status in a network segment relevant to a given task.

Unlike multi-task learning for the classification, the mutual relatedness is harder to assess in time series, as they are often a subject of time shifting, distortion and re-scaling effects. To address this problem, we first use the dynamic time warping distance measure, it helps identify similar series and group them together. Then we develop an alternative approach which attempts to infer a latent representation common to all observed time series. Such a common profile is used to align all series in Viterbi-like manner used in Markovian models. 

The contribution of this work can be summarized as follows:
\begin{enumerate}
\item We develop a multi-task learning approach to the joint learning for a set of time series. 
\item For a set or related time series, we design a support vector regression model derived from regularization-based techniques developed for the classification case.
\item For a given set of time series, we first use the dynamic time warping to measure their similarity; this helps identify clusters  of related time series.
\item We then propose another approach to address both time and scale warping of a set of time series, it infers their common latent profile and use this profile to align them among themselves. 
\item We test both methods on the travel demand prediction in Nancy public transport system and analyze advantages of each method. 
\end{enumerate}

The remainder of the paper is organized as follows. 
The problem of time series prediction in single and multi-task setting is introduced in Section II.
In Section III, we describe the regularization-based support vector regression for multi-task learning. 
Section III.A addresses the series relatedness and describes how the dynamic time warping distance can be used to identify similar time series. Section IV proposes an alternative approach to the series relatedness problem; it introduces the generative model for inferring a common latent representation of a set of time series and aligns them by both time and scale warping. Section V reports evaluation results of experiments with multi-task learning on a travel demand data collected in real public transport network and analyses the results. Finally Section VI concludes the paper.

\section{Time Series Prediction}\label{sec:timeseries}

Time series analysis and prediction theory have been intensively studied over last 40 years~\cite{sapankevych09,mills90}. 
Well known applications are financial market prediction, environment and weather forecasting, biology, control systems,  and many other applications involving non-linear processes~\cite{orfanidis}.

The goal of time series prediction is to estimate value $y$ at time $i$ based on past values $y_{i-1}$, $y_{i-2},\ldots$. It is common to limit the depth to the last $k$ values:
$y_{i}=f(y_{i-k},y_{i-k+1},...,y_{i-1}), i=k,\ldots,n,$
where $k\geq 1$ is the modeling depth (lag). Moreover, it can take into account a number of external characteristics $\bfv_i$ at time $i$ which may have impact on the series values:
\begin{equation}
y_{i}=f(\bfv_i,y_{i-k},y_{i-k+1},...,y_{i-1}), i=k,\ldots,n.
\end{equation}

The objective of time series prediction is to find a function $f(\bfx_i)$, where $\bfx_i=(\bfv_i,y_{i-k},y_{i-k+1},...,y_{i-1})$ such that ${\hat y}_i$, the predicted value of the time series at a future point in time, is consistent and minimizes a regularized fit function.


The underlying system models and time series data generating processes are generally complex and the models for these systems are usually not known in advance. Accurate and unbiased estimation of the time series data produced by these systems cannot always be achieved using linear techniques, and thus the estimation process requires more advanced time series prediction algorithms.

\subsection{SVM for Time Series Prediction}\label{ssec:svm}

The Support Vector Machine (SVM) is machine learning technique which has been successfully used for different tasks such as pattern recognition, object classification, and in the case of time series prediction and regression analysis. In Support Vector Regression (SVR), a function for a time series is estimated using observed data. Being a data-driven methodology, SVMs differ from more traditional time series prediction methodologies, like ARIMA or Kalman Filter~\cite{sapankevych09}. 

The underlying motivation for using SVMs for time series is the ability to make accurate prediction when the underlying system processes are typically nonlinear, non-stationary and not defined a-priori. In a number of applications, SVMs have been proven to outperform other non-linear techniques including neural-network based non-linear prediction techniques such as multi-layer perceptrons~\cite{lippi13,tan09,Muller97}.

Given a set of time series data $(\bfx_i,y_i)$, $i=1,\ldots,n$, a time series prediction method defines a function $f(\bfx)$ that will have an output equal to the predicted value for some prediction horizon. By using regression analysis, the prediction function for the linear regression is defined as 
$f(\bfx)= \bfw^T \bfx +b$.
If the data is not linear in its input space, the data $\bfx$ are mapped into a higher dimension space, via a kernel function $\phi(\bfx)$, then perform a linear regression in the higher dimensional feature space~\cite{Muller97,Ruping01}:
$$f(\bfx)= \bfw^T \phi(\bfx) +b.$$ 
The goal is therefore to find optimal weights $\bfw$ and threshold $b$, as well as to define the criteria for finding an optimal set of weights. 

Given training data $(\bfx_i,y_i), i=1,\ldots,n$, $\bfx_i \in {\cal R}^d$, $y_i \in {\cal R}$, 
SVM first maps input vectors $\bfx$ onto the feature space $\Phi, \phi(\bfx) \in \Phi$  and then approximates the regression by a linear function $f(\bfx) =\bfw^T \phi(x) + b$. This is obtained by solving the following optimization problem in the $\varepsilon$-insensitive tube~\cite{smola04}:
\begin{equation}
\begin{array}{ll}
{\rm min}     & \frac{1}{2}||\bfw||^2 +C \sum_{i=1}^n (\xi_i+\xi^*_i)  \\
{\rm s.t.} & \left\{  
  \begin{array}{l} 
       y_i -(\bfw^T\phi(\bfx_i) -b) \leq \varepsilon +\xi_i \\
       (\bfw^T\phi(\bfx_i)+b)-y_i \leq \varepsilon+\xi_i^*,i=1,\ldots,n \\
       \xi_i^*\geq 0, \xi_i \geq 0,i =1,\ldots,n\\
  \end{array} \right.
\end{array} 
\label{eq:svr}
\end{equation}
\noindent
where $\xi_i, \xi_i^*, i=1,\ldots,n$ are slack variables, measuring the deviation from $\varepsilon$-insensitive tube, $C$ is the regularization parameter.

\section{Multi-task SVM Regression}\label{sec:mtl-svr}

Suppose we dispose  a set of $m$ related time series, $m>1$. Training samples are represented as a set
$\{(\bfx_{ri},y_{ri}), r=1,\ldots,m, i=1,\ldots,n\}$. We denote the indices from series $r$ by $T_r=\{i_{r1},\ldots,i_{rn} \}, r=1,\ldots,m.$~\footnote{Without loss of generality, we make assumption of the same length $n$ for all series.}

Making an assumption of relatedness between the tasks serves us to design the multi–task learning method. We assume that the tasks are related in a way that the true models are all close to some common model $\bfw$~\cite{evgeniou2004} and every task model $\bfw_r$ can be written as $\bfw + \bfw_r$, $r=1,\ldots,m$, where the individual vectors $\bfw_r$ are small when the tasks are similar to each other. 

We adapt the regularization approach~\cite{evgeniou2004}, according to which vectors of each task $\bfx_i, i\in T_r$ are mapped into two different spaces. One is a space $\Phi$ common to all tasks, $\phi(\bfx_i) \in \Phi$; another is a correction space $\Phi_r$, specific to every task $r$, $\phi_r(\bfx_i)\in \Phi_r$. 

Individual task information is accounted in the slack variables, defined as follows:
\begin{equation}
\begin{tabular}{lcl}
$\xi_i$   & =& $\bfw_r^T \phi(\bfx_i)+ b_r,     i \in T_r, r=1,\ldots,m$ \\
$\xi_i^*$ & =& ${\bfw_r}^T \phi(\bfx_i)+ b^*_r, i \in T_r, r=1,\ldots,m.$ \\
\end{tabular}
\end{equation}

All slack variables are non-negative, $\xi_r(\bfx_i),\xi^*_r(\bfx_i) \geq 0,r=1,\ldots,m$. Thus samples maped in the correcting space have to lie on one side of the corresponding correcting function. The correcting function also has to pass through some points with slack variables being zero.
 
The goal of the multi-task version of SVR is to estimate $m$ regression models, one model per task. 
Multi-task learning SVR incorporates individual task information into estimated regression functions through the slack variables. We specify the following parameterized functions for $m$ regression models:
\begin{equation}
f (\bfx) +f_r (\bfx) =\bfw^T \phi(\bfx)+b +\bfw_r^T \phi_r(\bfx) +b_r, r=1,\ldots,m,
\label{model-mtl-svr}
\end{equation}
where $f (\bfx)=\bfw^T \phi(\bfx)+b$ is the common estimation function and $f_r (\bfx)=\bfw_r^T \phi_r(\bfx)+b_r$ is a correction function for series $r$, $r=1,\ldots,m$. Therefore the MTL SVR formulation solves the following optimization problem:
\begin{equation}
\begin{array}{ll}
{\rm min} &\frac{1}{2}||\bfw||^2+\mu\sum_{i=1}^t ||\bfw_r||^2+C\sum_{i=1}^n(\xi^r_i+\xi^{r*}_i)\\ 
{\rm s.t.} & \left\{ 
    \begin{array}{l}
	y^r_i -(\bfw^T \phi(\bfx_i)+b+{\bfw_r}^T\phi_r(\bfx_i)+b_r) \leq \varepsilon +\xi^{r}_i \\ 
	(\bfw^T\phi(\bfx_i)+b +{\bfw_r}^T \phi_r(\bfx_i)+b_r) -y^r_i \leq \varepsilon +\xi^{r*}_i \\	 
	\xi^{r*}_i \geq 0, \xi^r_i \geq 0, i=1,\ldots,n, r=1,\ldots,m. \\ 
    \end{array} \right.
\end{array}
\label{eq:mtl-svr}
\end{equation}
where $\mu$ is a regularization parameter for the individual correction functions. Using the dual optimization technique (similar to standard SVM), the dual form of the above optimization problem is as follows:
\begin{equation}
\begin{array}{ll}
{\rm max} & -\varepsilon \sum_{i=1}^n (\al_i^* +\al_i) + \sum_{i=1}^n (\al_i^* -\al_i) y_i  \\
               & -\frac{1}{2} \sum_{i,j=1}^n (\al_i^* -\al_i)(\al_j^*-\al_j)\phi(\bfx_i)^T \phi(\bfx_j) \\
	       & -\frac{1}{2\mu} \sum_{r=1}^t \sum_{i,j \in T_r} (\al_i^* -\al_i) (\al_j^* -\al_j){\phi_r(\bfx_i)}^T\phi_r(\bfx_j) \\
{\rm s.t.}& \left\{ \begin{array}{ll}
		              \sum_{i \in T_r} (\al_i^*-\al_i)=0, r=1,\ldots,m,\\
		              0 \leq \al_i, \al_i^* \leq C, i=1,\ldots,n. \\
		            \end{array} \right.
\end{array}
\label{eq:mlr-dual}
\end{equation}

Then vectors $\bfw$, $\bfw_r$ can be expressed in terms of training samples:
\begin{equation}
\begin{array}{rll}
\bfw   &=& \sum_{i=1}^n (\al_i^* -\al_i)\phi(\bfx_i),\\
\bfw_r &=& \frac{1}{\mu} \sum_{i \in T_r} (\al_i^* -\al_i) \phi_r(\bfx_i), r=1,\ldots,m.\\
\end{array}
\label{eq:kkt}
\end{equation}

Note that in addition to usual parameters of individual SVR, $C$ and $\varepsilon$,  
the MTL extension requires also to tune parameter $\mu$.

\subsection{Multiple Series Relatedness}\label{ssec:dtw}

In the previous section, multi-task learning was assumed to be applied to a set of mutually related time series. 
Ensuring the task relatedness is crucial; applying the multi-task learning to unrelated series may have a negative impact on the prediction quality.

A pair of time series is said to be mutually related if they behave similarly and show up the same tendencies.
To address the relatedness between time series, we first consider conventional measures of similarity 
which is of fundamental importance for time series analysis~\cite{Fu2011}.
The straightforward approach to measure the similarity between two time series is to evaluate the Euclidean
distance on their normalized representation.
However, Euclidean distance ignores any shifting and distortion in time series and has been found unsuitable in different domains~\cite{dtw08}. 

Besides the Euclidean-based distance measure, the time-warping distance has become one of the most popular and field-tested similarity measures. The {\it dynamic time warping} (DTW) distance is efficient as the time-series similarity measure for minimizing the effects of shifting and distortion in time; it allows elastic transformation of time series in order to detect similar shapes with different phases. Given two time series, DTW algorithm yields the optimal solution in the $O(n^2)$ time.

Since many time series mining algorithms use similarity search as a core subroutine, 
To avoid that the time taken for DTW similarity becames a bottleneck, multiple efforts have been undertaken to improve the complexity of the DTW distance algorithms. 
In recent work~\cite{rakthanmanon2012}, a novel DTW similarity algorithm for time series is proposed in order to search and mine massive time series. It permits to search with DTW similarity faster than with the Euclidean distance algorithm. For such problems as motif discovery and time series clustering, it scales up to very large datasets accounting for billions of time series.


Below we use the DTW similarity to identify groups of similar time series and then apply the multi-task learning to each of the identified clusters. Algorithm 1 below summarizes the multi-task learning on a given set of time series.
 
\begin{algorithm}
\begin{algorithmic}[1]
\REQUIRE Set of time series $\bfy_1,\ldots,\bfy_m$, $\bfy_r=(y_{r1},\ldots,y_{rn})$, $r=1,\ldots,m$
\REQUIRE Number of clusters $L$
\REQUIRE Observations $(\bfv_r,\ldots,\bfv_{rn})$, including weather, calendar and traffic data, for every series $\bfy_r$. 
\FOR {every pair $(\bfy_r,\bfy_t), 1\leq r<t\leq m$}
  \STATE Measure the similarity between series $\bfy_r$ and $\bfy_t$ by the dynamic time warping (DTW) distance. 
  \STATE If series are too long (more 10,000 elements), uniformly sample series before applying the DTW algorithm.
\ENDFOR
\STATE Apply agglomerative clustering algorithm to form $L$ clusters $G_1,\ldots,G_L$ of series using DTW distance as metric. Any series $\bfy_r$ belongs to exactly one group.
\FOR {each group $G$ of relates tasks}
 \STATE Apply the multi-task learning described in Section~\ref{sec:mtl-svr} 
\ENDFOR
\RETURN Multi-task models for all groups and tasks
\end{algorithmic}
\caption{Multi-task learning for related and unrelated time series.} 
\label{alg1}
\end{algorithm}

\section{Time series common profile}\label{sec:cpm}

In the previous section, we grouped time series by using their DTW distance as similarity metric. This helps identify similar tasks but does not guarantee their relatedness. 

In this section, we make a step further in the analysis of time series relatedness, by taking into account both time warping and scale warping when aligning two or more time series.
Our approach is based on an assumption that all series in the set are noisy and rescaled replicates of one common and latent representation~\cite{listgarten04}. To leverage the information contained in such noisy replicates, such a common representation can be inferred and used to align the series among themselves in an appropriate way.  




The latent trace is modeled as an underlying, noiseless representation of the set of observed noisy time series. 
Output time series are generated from this model by moving through a sequence of hidden states and emitting an observable value at each step, as in Hidden Markov Machines (HMMs). 
To account for changes in the value amplitude within and across replicates, the latent time states are augmented by a set of scale states, which control how the emission signal will be scaled relative to the value of the latent trace.
The latent trace is learned in unsupervised manner, as well as the transition probabilities controlling the Markovian evolution of the scale and time states and the overall noise level observed data. 

The Continuous Profile Model (CPM)~\cite{listgarten04} is a generative model for a set of $m$ time series, $\bfy_1, \bfy_2, ...,\bfy_m$.
Time series in the set are noisy replicas of a common and latent representation, defined as a trace $\bfz=(z_1,\ldots,z_M)$.
Any given time series $\bfy_r$ is a sub-sampled version of $\bfz$ after applying local scale transformations.
Size $M$ of $\bfz$ is larger than time series size $n$ so that any experimental data could be mapped precisely to the correct underlying trace point. 
In practice, $M =\lambda n$ with $1 < \lambda< 3$. 

The resolution of the latent trace $\bfz$ is higher with respect to the observed time series, so the experimental time can be made effectively to speed up or slow down by advancing along the latent trace in larger or smaller jumps. The sub-sampling and local scaling used during the generation of each observed time series are determined by a sequence of hidden state variables. The state sequence for series $\bfy_r$ is denoted ${\pi^r}$. Each state in the sequence maps to a pair (time state, scale state), $\pi^r_i \rightarrow \{\tau_i^r, \psi_i^r \}$, where time state $\tau_i^r \in 
(\tau_1,\ldots,\tau_M$), and scale states $\psi_i^r \in (\psi_1,\ldots,\psi_Q$), $Q$ is the number of scale states.

The probability of emitting observed value $y_i^r$ in state $\pi_i^r$ is defined by the emission probability distribution
$A_{\pi_i^r} (y^r_i |\bfz) \equiv p(y_i^r |\pi_i^r, \bfz, \sigma, u^r )$ which is defined by probability distribution 
${\cal N} (y_i^r; z_{\tau_i^r} \psi_i^r u^r, \sigma)$, where $\sigma$ is the noise level of the observed data, ${\cal N} (a;b,c)$ denotes a Gaussian probability density for $a$ with mean $b$ and standard deviation $c$. The $u_r$ are real-valued scale parameters, one per observed time series, that correct for any overall scale difference between time series $r, r=1,\ldots,m$, and the latent trace $\bfz$.

The transitions between time states are independent from transitions between scale states, so 
$p(\pi_i^r |\pi^r_{i-1} ) = p(\psi_i |\psi_{i-1}) p^r(\tau_i |\tau_{i-1})$. To enforce that time must move
forward, any time state can jump ahead no more than $J_\tau$ states. 
Similarly, scale state transitions are allowed between neighboring scale states only, this disallows arbitrary jumps on the local scale. These constraints keep the number of legal transitions to a tractable computational size. Each observed time series $\bfy^r$ has its own time transition probability distribution to account for experiment-specific patterns. The time transition probability distribution for $\bfy^r$ is given by multinomial distribution as follows
\begin{equation}
p^r(\tau_i = a|\tau_{i-1} = b) = \left\{
	\begin{array}{ll}
		d_l^r, & {\rm if\ } a - b = l, l=1,\ldots,J_\tau, \\
	    0,     & {\rm otherwise.} \\
	 \end{array} \right.
	 \label{eq:timetrans}
\end{equation}
where $\sum_{i=1}^{J_\tau} d^r_i = 1$. The time scale probability transition for all series is given by 
\begin{equation}
p(\psi_i = a|\psi_{i-1}=b) = 
      \left\{ 
		\begin{array}{ll}
			s_0, & {\rm if\ } D(a,b) = 0,     \\
			s_1, & {\rm if\ } D(a,b)=\pm 1,   \\
			0,     & {\rm otherwise,} \\
		\end{array} \right.  
\label{eq:timescale}
\end{equation}
where $D(a,b)=\pm 1$ means that $a$ is one scale state larger (+1) or smaller (-1) than $b$, and $D(a,b)=0$ means that $a=b$ and the scale state does not change. The distribution $J_\tau$ are constrained by $2 s_1 + s_0 = 1$.

$J_\tau$ determines the maximum allowable speedup of one portion of a time series relative to another portion, within the same series or across different series. However, the length of time for which any series can move so rapidly is constrained by the length of the latent trace; thus the maximum overall ratio in speeds achievable by the model between any two entire time series is given by $min(J_\tau,\lambda)$.

After training, one may examine either the latent trace or the alignment of each observed time series to the latent trace. Such alignments can be achieved by several methods, including use of the Viterbi algorithm to find the highest likelihood path through the hidden states~\cite{baum1970}, or sampling from the posterior over hidden state sequences.  Viterbi alignment is found to work well in our experiments. 

Training the CPM with the EM algorithm is similar to HHMs with Baum-Welch algorithm. The $E$-Step is computed exactly using the Forward-Backward algorithm, which provides the posterior probability over states for each time point of every observed time series 
and the pairwise state posteriors. 
The $M$-Step consists of a series of analytical updates to the various parameters, see~\cite{listgarten04} for more detail.

\begin{figure}[ht]
\centering{
\includegraphics[width=8cm]{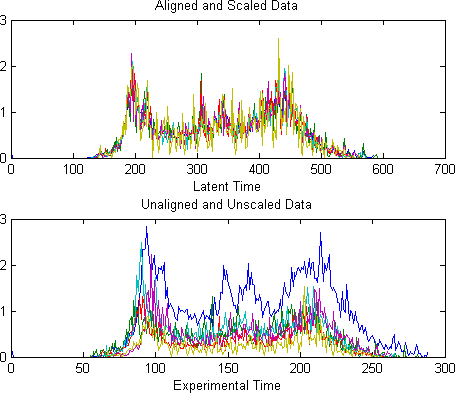}
}
\caption{Time and scale alignment of 5 travel demand time series.} 
\label{fig:alignment}
\end{figure}

Figure~\ref{fig:alignment} shows an example of alignment for five travel demand times series. 
 
\section{Evaluation}\label{sec:evaluation}

We run experiments to test learning methods described in previous sections on the task of travel demand prediction in Nancy (France) public transport network. Two alternative approaches to model a set time series are tested in different settings and scenarios. One is a single-task learning approach where a separate support vector regression model is estimated for each task independently. 
Another is the multi-task learning approach where one common and a number of individual correction models are jointly learned to estimate related regression models. A set of time series representing the travel demand are analyzed and pre-processed. In one instantiation, they are clustered by using the DTW distance as described in Algorithm 1. In another, a latent profile common to all the series is inferred and then all series are aligned to this common profile, as described in Section IV. 



Public is GDP time series include annual per capita GDP time series for several countries. One interest lies in studying the "periodic" behaviour of such series in connection with understanding business cycles. Another lies in forecasting turning points. Another lies in comparisons of characteristics of the series across national economies.

We dispose the following dataset to evaluate different prediction models:
\begin{itemize}
\item Historical data of the travel demand, which accounts for the total of 5,29 M passenger boarding events registered at 1,067 stops and 28 tram and bus lines in the city network; the dataset covers the period of 90 days from March to May 2012, including 13 weekends, 2 weeks of school vacations and three public holidays. Each boarding transaction includes the route, stop, direction and the timestamp of the event. Time series describing the travel demand are then generated by the aggregation of boarding timestamps at a regular time period. In the tests, two different periods are used to produce the time series, of 10 and 60 minutes.
\item Weather data include the air temperature (minimum and maximum), air humidity, precipitation, wind speed, etc. 
All data are registered hourly~\footnote{Available from weather station Nancy-Ochey at {\rm http://www.meteociel.fr/temps-reel/ obs\_villes.php?code2=7181}.}. 
\item Calendar data, including the binary tags for weekends, school vacations and public holidays.
\item Traffic information on the network segment relevant to a given task. The average bus adherence delay accumulated between timestamps $i-1$ and $i$ for each series is used.
\end{itemize}

\subsubsection{Protocol and experimental setup}
\label{ssec:protocol}

We compare the performance of different methods for time series by using the standard measure of Mean Square Error (MSE). It is defined as follows
\begin{equation}
E_{MSE} = \frac{1}{n} \sum\limits_{i=1}^{n} \left(y_i-\hat{y}_i \right)^2,
\end{equation}
where $y_i$ is the true value at time point $i$ in the test set while $\hat{y}_i$ is the prediction.

\begin{figure}[ht]
\centering{
\includegraphics[width=10cm]{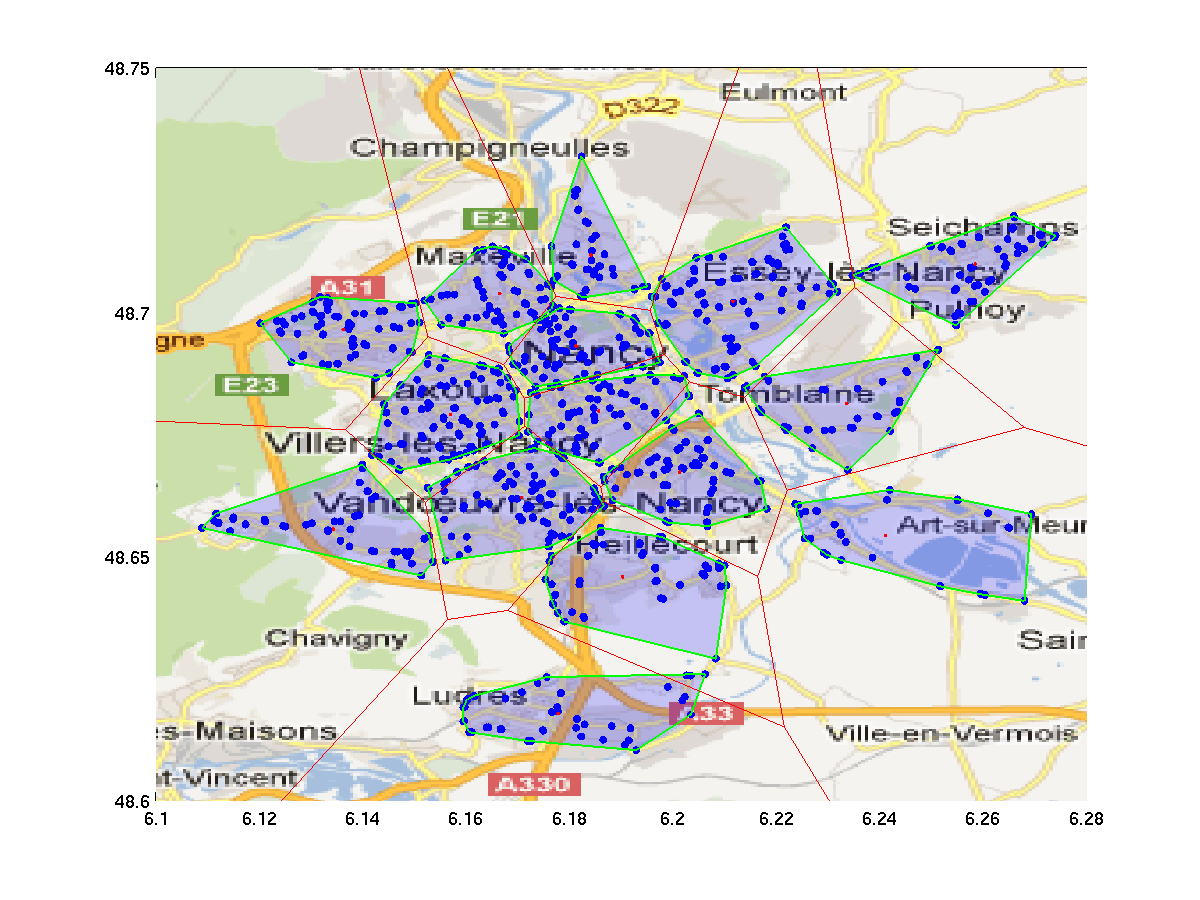}}
\caption{Nancy city zoning by travel demand.} 
\label{fig:nancy-zone}
\end{figure}

Experiments for the travel demand prediction are conducted on two levels of the public transport network; one on city zones, another for individual stops. For the {\it zone demand} experiment, a preliminary analysis of the travel dataset has been done, with the goal to extract all travel patterns required for the inference of origin-destination (OD) matrices. The city network is then segmented into zones, where one zone is defined as a compact city region with stops showing up a high similarity of their travel patterns. Figure~\ref{fig:nancy-zone} demonstrates the segmentation of Nancy city network into 15 zones, where blue points refer to the bus and tram stops and each zone is shown as a convex hull built around geographical locations of all stops included in the zone. For this case, we generate 15 times series describing the travel demand in all zones. 

For the {\it stop demand} experiment, one tram and four bus lines have been selected.
For each line, a representation set of 10 stops has been selected to produce the set of travel demand time series. 

The evaluation prototype has been implemented in MATLAB as extension of MALSAR toolkit for multi-task learning via structured regularization~\cite{zhou2012}.

\begin{figure}[ht]
\centering{
\includegraphics[width=5cm]{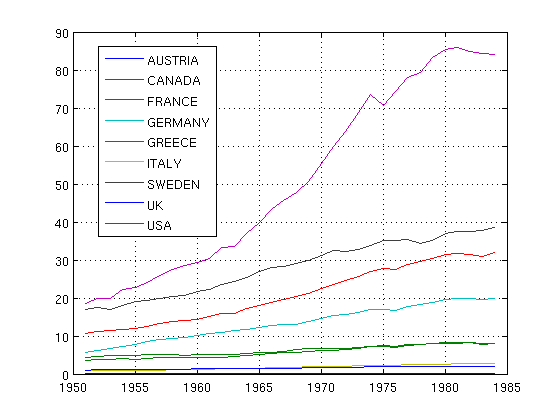}
\includegraphics[width=7cm]{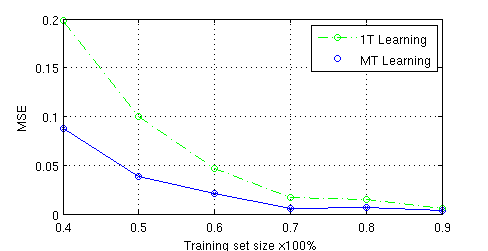}}
\caption{GDP per capita: a) 9 series, b) 1T and MT learing.} 
\label{fig:gdp}
\end{figure}

In all experiments, all time series are split into training and testing sets in the same way. The training set includes starting $P$\% items (days); the testing set includes remaining $100-P$\%, where $P$ varies from 40 to 90.  
 
For each of two experiments, we evaluate three different learning approaches in different scenarios:
\begin{enumerate}
\item We compare the prediction performance of the three following methods: 
\begin{itemize}
  \item Single-task (ST) learning, where individual SVR models are learned and tested independently; the average of their MSE values is reported;
  \item Multi-task (MT) learning, where one common SVR model and one correction model per time series are learned; the series are clustered by their DTW distance, as described in Section~\ref{sec:mtl-svr}.  
  \item MT learning with CPM alignment, where the time series get aligned to their common profile, as described in Section~\ref{sec:cpm}.
  \end{itemize}	
\item We run experiments with two different demand aggregation periods, of 10 minutes and 1 hour.
\end{enumerate}

All time series are normalized and scaled to the logarithmic space. In SVR, the linear kernel function $\phi(\bfx)=\bfx$ is preferred to the Gaussian one. The depth $k=2$ for the past values (lag) was determined by cross-validation and remains the same in all experiments. Parameters $C$ and $\varepsilon$ are automatically tuned as in the standard SVM regression~\cite{smola04}, the optimal value of parameter $\mu$ is determined by re-sampling. 

Algorithm 1 for clustering time series by their DTW distance takes as input the number of clusters $L$. 
In the zone experiment, where we produce 15 time series, the number of clusters to test is between 2 and 15.
In the stop experiment, 10 time series represent any selected line, and the number of clusters is between 2 and 10.
The optimal number of clusters is detected by the cross-validation, the best results are reported for $L$ between 3 and 6 in the stop experiment and 4 or 5 in zone experiment.

For the CPM inference and alignment, seven latent traces $\bfz_1,\ldots,\bfz_7$ are inferred from the daily times series in the training set, one for each day of week. At the test time, a daily time series $\bfy_i$ as well as relevant observations $\bfv_i$ get aligned to the same day trace $\bfz$ using the Viterbi algorithm. The size $M$ of the latent for series of size $n$ is $\lambda n$ where $\lambda=2.5$. The scaling space size $Q$ is limited to 5 states, they are evenly spaced on logarithmic scale. The MSE values are then averaged over 7 day-specific experiments. Finally, the number of iterations for EM algorithm was limited to 20.
 
In all figures presented below, the whisker plots are used to show the mean, variance and outliers of MSE values.

\subsubsection{Stop experiments}\label{ssec:stop}

\begin{figure}[ht]
\centering{
\includegraphics[width=8cm]{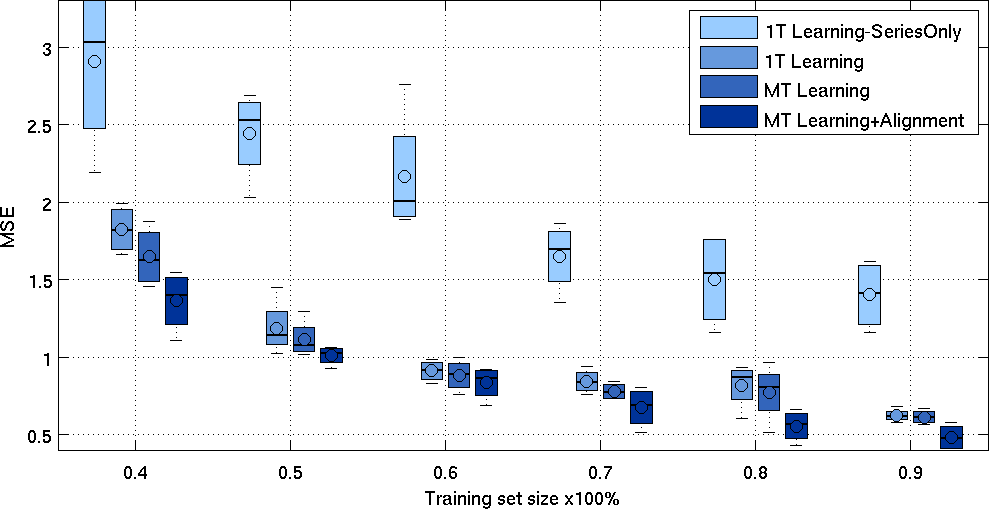}
}
\caption{Stop demand prediction for 1 hour period.
} 
\label{fig:stop}
\end{figure}

Figure~\ref{fig:stop} reports evaluation results of single- and multi-task learning for the stop experiment with 1 hour period. In addition to the three referred methods, it also includes MSE values for the single-task learning without external observations $\bfv_i$ (weather, calendar and traffic features). As one can see, taking external observations into account reduces the error values by 40\%-60\%; this measures the impact of external observations on the predictive behavior of support vector regression models
\footnote{Analysis of individual contribution of external observations is beyond the scope of this study.}.
The multi-task learning with DTW outperforms the single-task learning by 2\%-7\%, multi-task learning with CPM further reduces the error by 3\%-12\%.

\begin{figure}[ht]
\centering{
\includegraphics[width=8cm]{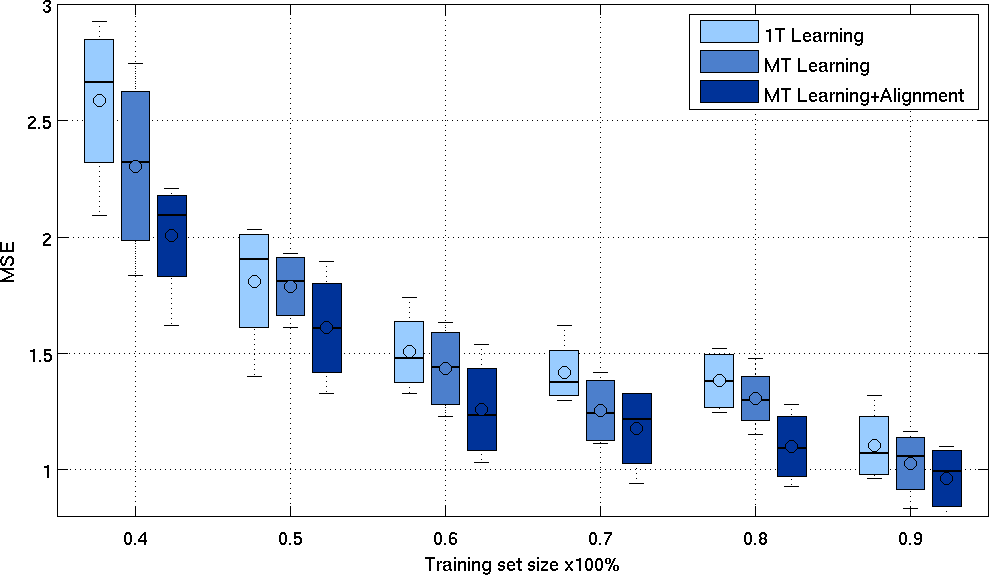}}
\caption{Stop demand prediction for 10 minute period.} 
\label{fig:stop15}
\end{figure}

Figure~\ref{fig:stop15} reports the error values for the stop experiment, this time with 10 minute period. Shorter period of data aggregation leads to a higher variance in time series; this results in higher error values. Both multi-task learning methods improve the predictive performance with respect to single-task learning, with multi-task learning with CPM being the best.

\subsubsection{Zone experiments}\label{ssec:zone}

Figure~\ref{fig:stop} reports the evaluation results of the referred methods for the zone experiment in 0-delay scenario, with 1 hour period. The multi-task learning with DTW outperforms the single-task learning by 0\%-9\%, multi-task learning with CPM reduces the error by 3\%-12\%.

Figure~\ref{fig:stop} reports results for the same experiment in 1-delay scenario. Unavailability of the last value in series makes the prediction harder for all methods, both in terms of the mean and the variance. The multi-task with DTW plays the equal game with the single-task learning or outperforms it in some cases. 

\begin{figure}[ht]
\centering{
\includegraphics[width=6cm]{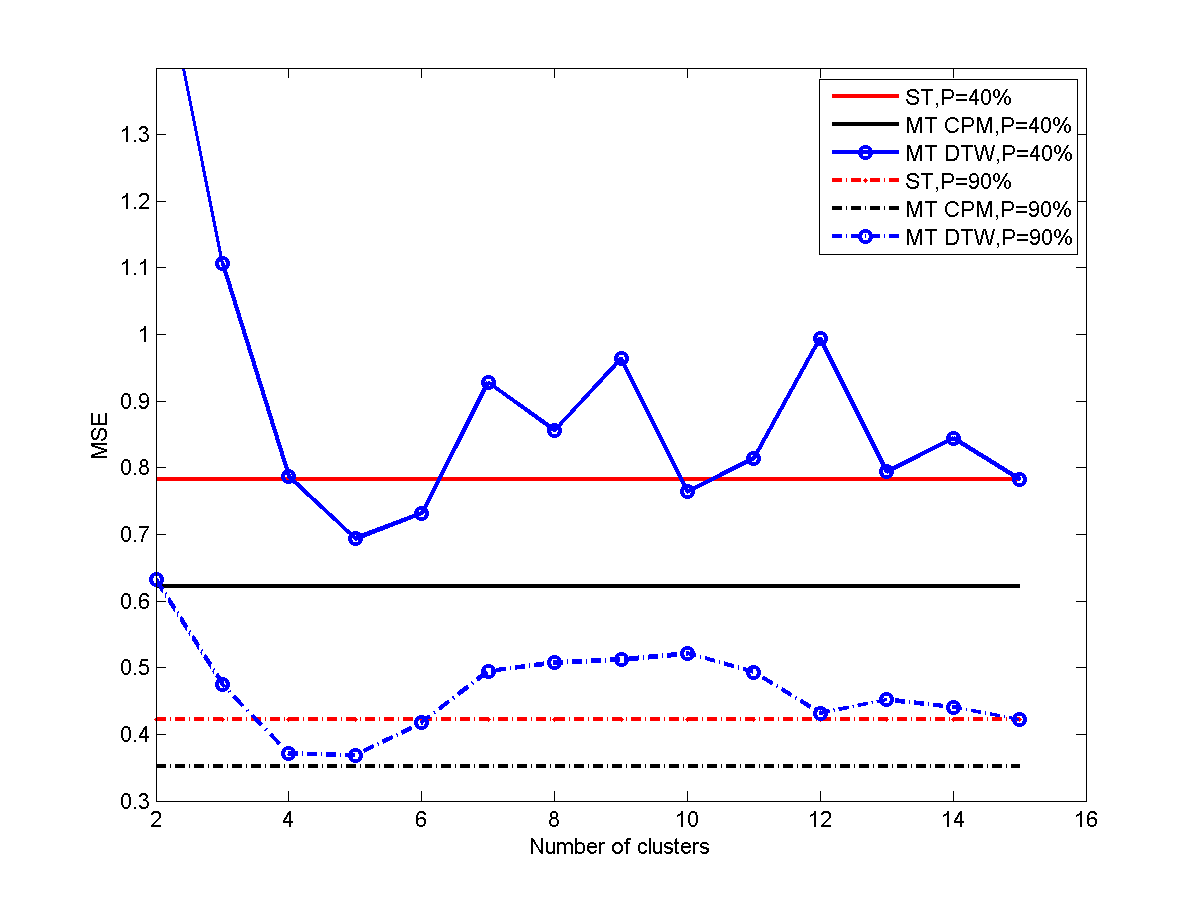}
\includegraphics[width=6cm]{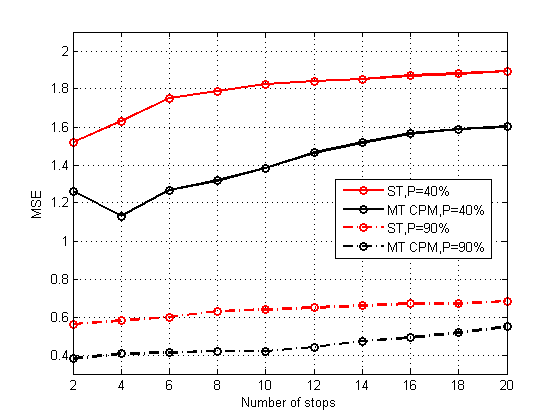}
}
\caption{a) MT with DTW for different number of clusters.
b) MT with CPM for different numbers of stops.
} 
\label{fig:dtw}
\end{figure}

In this experiment, multi-task learning with DTW performs better than the single-task learning for few values of $L$; in other cases, it often shows a worse performance (see Figure~\ref{fig:dtw}.a for cases $P$=40\% and 90\%). The multi-task learning with CPM resists better to the data delay, it outperforms the single-task learning by 3\%-16\%.

\begin{figure}[ht]
\centering{
\includegraphics[width=8cm]{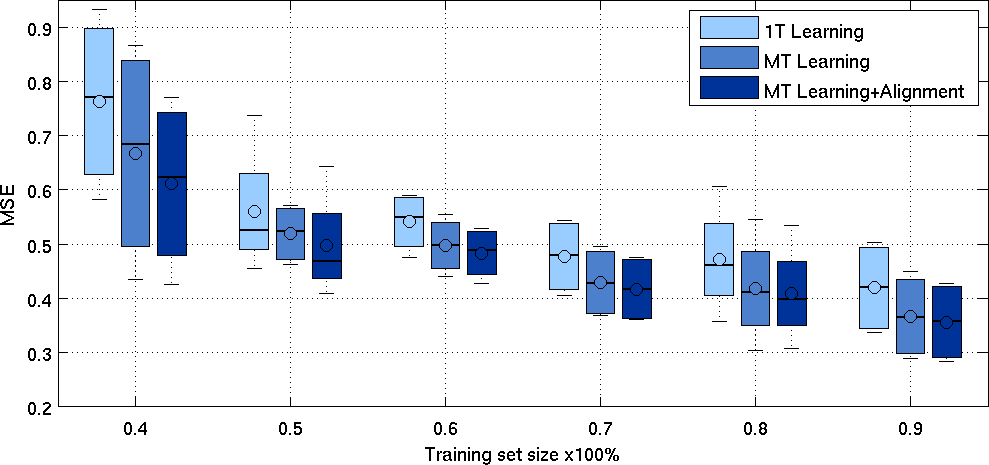}}
\caption{Zone demand prediction.} 
\label{fig:zone}
\end{figure}

\subsubsection{Discussion} 
\label{ssec:discussion}

Analysis of the evaluation results in the previous subsections suggests that multi-task learning can offer a comfortable performance improvement over single-task setting. However, unlike the multi-task learning for classification, the task relatedness raises a particular concern. A simple normalization is insufficient and advanced techniques are required to leverage the information shared by different time series.

Between two methods presented in Sections III and IV, the continuous profile model (CPM) gives an advantage over the DTW distance based clustering. Considered as a straightforward choice to identify related time series, Algorithm 1 is sensible to the way the times series get clustered. Moreover, DTW-based alignment does not guarantee the relatedness of selected tasks.  

Time series alignment with CPM seems to solve the problem, the method appears more stable, it yields a performance improvement in a majority of experiments and scenarios. The number of time series we tested in both zone and stop experiments is modest and do not raise particular issues when inferring daily latent traces, even for 10 minute period. In Figure~\ref{fig:dtw}.b we relax the number of stops in the stop experiment. Instead of 10 representative stops,  we pick up randomly 2 to 20 stops, and compare multi-task CPM to single-task learning, for $P$=40\% and 90\%.

Computationally, multi-task DTW and CPM methods behave similarly, due to the straightforward dynamic programming implementation of the DTW algorithm and a reasonable convergence of the EM algorithm when inferring 7 daily latent traces and Viterbi-based alignment of the observed time series.

\section{Conclusion} \label{sec:conclusion}

In a public transport network, an accurate estimation of the number of travelers entering the network at any period of time is crucial for efficient planning of transport services. The estimation model is learned from historical data and relevant observations. For a set of time series describing the demand, the single-task learning can be further turned into by multi-task learning in order to leverage the information shared by different time series. In this paper, we describe the regularization-based support vector regression for multi-task learning and propose two ways of addressing the series relatedness. One way is to cluster times series by their dynamic time warping distance. Another way is to infer a latent representation common to all time series and then align the series by this common representation. We apply the proposed methods on the travel demand prediction in Nancy public transport network and report the evaluation results for single- and multi-task learning methods.


\bibliographystyle{plain}
\bibliography{localBiblio}

\end{document}